\documentclass[letterpaper, 10pt, conference]{ieeeconf}      % Use this line for a4 paper
\IEEEoverridecommandlockouts                              % This command is only 
\usepackage{subcaption}
\usepackage{amsmath,amsfonts,amssymb}
\usepackage{bm}
\usepackage{amssymb,amssymb,calligra,calrsfs}
\usepackage{soul}
\usepackage{algorithm2e}
\RestyleAlgo{ruled}
\usepackage{array}
\usepackage[caption=false,font=normalsize,labelfont=sf,textfont=sf]{subfig}
\usepackage{textcomp}
\usepackage{stfloats}
\usepackage{url}
\usepackage{verbatim}
\usepackage{graphicx}
\usepackage{cite}
\usepackage{authblk}
\usepackage{xcolor}
\usepackage{booktabs}
\usepackage{multirow}

\newcommand{\norm}[1]{\lvert {#1} \rvert}

\title{\LARGE \bf Outlier-Robust Nonlinear Moving Horizon Estimation using Adaptive Loss Functions}

\author{Nestor Deniz$^{1,2}$, Guido Sánchez$^{1}$, Fernando Auat Cheein$^{2}$ and Leonardo Giovanini$^{1}$% <-this % stops a space
\thanks{*This work was supported by Consejo Nacional de Investigaciones Científicas y Técnicas (CONICET) from Argentina}% <-this % stops a space
\thanks{$^{1}$ Instituto de Investigaci\'on en Senales, Sistemas e Inteligencia Computacional, s\i nc(\textit{i}), UNL-CONICET, Santa Fe, Argentina (e-mail: ndeniz@sinc.unl.edu.ar).}
\thanks{$^{2}$ Harper Adams University, Engineer Department, Edgmond, Newport TF10 8NB (e-mail: ndeniz@harper-adams.ac.uk).}
}

\begin{document}

\maketitle

\thispagestyle{empty}
\pagestyle{empty}

\begin{abstract}
In this work, we propose an adaptive robust loss function framework for MHE, integrating an adaptive robust loss function to reduce the impact of outliers with a regularization term that avoids naive solutions. The proposed approach prioritizes the fitting of uncontaminated data and downweights the contaminated ones. A tuning parameter is incorporated into the framework to control the shape of the loss function for adjusting the estimator's robustness to outliers. The simulation results demonstrate that adaptation occurs in just a few iterations, whereas the traditional behaviour $\mathrm{L_2}$ predominates when the measurements are free of outliers.
\end{abstract}

\section{Introduction}
State estimation is essential for feedback control, system monitoring, and optimization, as only noisy measurements are typically available from the system. The challenge increases when measurements are affected by sporadic noise that does not follow a specific distribution. Such deviations often result in outliers, individual measurements that differ significantly from expected values, which can severely degrade the accuracy of estimation. Several authors have addressed this challenge using different techniques in which variants of the Kalman Filter (KF) have been extensively applied.

The works \cite{Rahimnejad2024178552, Gao2025, Bao2025, Zhang20257955, Biswal2025, Lee2023132766, Si2023} use the Kalman Filter (\emph{KF}) as a central tool to perform state estimation in general, and some variants for measurements with outliers. These variants include KFs lattices, kernel density estimation, geometric approaches, and maximum correntropy with Student's $t$ and robust kernels. Although the KF has proven to be an excellent filtering tool since its appearance in $1960$, it does not allow the incorporation of constraints naturally, leading to inaccurate estimates when the system is complex \cite{Haseltine2005, Deniz2023678}.

Inspired by the success of moving-horizon methods, moving-horizon estimation (MHE) has received growing attention. MHE is formulated as an optimization problem; thus, the inclusion of constraints and multirate sampling is naturally supported. MHE solves a finite-horizon state estimation problem at each sampling time. When a new measurement becomes available, the old one is discarded from the estimation window. Information outside of the estimation window is included in the objective function through an additional term called \textbf{arrival cost}. A good approximation of the arrival cost allows one to reduce the size of the estimation window and to have a performance comparable to that of the full-information estimator. Results on robust stability and convergence properties have been obtained in recent years, advancing from idealistic assumptions (observability and no disturbances) to realistic situations (detectability and bounded disturbances). 

In this regard, robust stage-cost functions have been proposed instead of the standard quadratic stage-cost to address the challenges of state estimation under heavy noise conditions and measurements affected by outliers. For example, the method in \cite{ALESSANDRI201685} solves $N+2$ parallel optimization problems per step to identify the measurement subset with the lowest cost, rejecting the rest to preserve robustness. The authors in \cite{PALMA202014636} combine the \textbf{Least Median of Squares} (LMS) criterion with the \textbf{Rauch–Tung–Striebel} (RTS) smoothing for outlier mitigation in marine robotics, improving temporal consistency. The MH-RAPS estimator in \cite{https://doi.org/10.1002/acs.3055} incorporates a binary measurement selection vector in a constrained MHE formulation, ensuring a Fisher information matrix bound for risk-aware outlier exclusion. Finally, in \cite{10156391}, the Generalized MHE (GMHE) replaces the \emph{KL} divergence with the $\beta$-divergence in the loss function, resulting in a posterior estimation more tolerant to outliers via maximum a posteriori optimization. These methods improve robustness against outliers and increase computational burden.

In our previous work \cite{deniz2025robust}, we proposed the use of a robust stage-cost function defined as the square of the derivative of the general adaptive robust loss function presented in \cite{Barron_2019_CVPR}. We found that this new adaptive loss function shows a similar behaviour to the original one, but using the $[1,\,2)$ interval of the shape-controlling parameter $\alpha$ instead of an infinite interval. In addition, it also avoids the singularity present in the original function. We also found that for $\alpha\approx 1.5$, the MHE achieves a good performance in the presence of outliers. In this work, we extend this idea by incorporating the online adaptation of $\alpha$ within an iterative scheme in which states and $\alpha$ are computed iteratively until a stop condition is reached.

The paper is structured as follows. Section II introduces the kind of system and the adaptive robust loss function that we use to solve the outlier problem. In the same Section, we formulate the proposed estimator first in continuous time and its discrete counterpart, which is the one that is solved. Robust stability is also briefly analysed. In Section III, the performance of the proposed estimator is evaluated and compared with our previous method \cite{deniz2025robust} and an estimator that searches for the value of the shape parameter through grid search. Section IV concludes and discusses future directions.

\section{Problem Statement}
\subsection{Preliminaries}
Let us consider the state estimation problem for a discrete-time nonlinear system given by
\begin{equation}        \label{eq: eq_nonlinsys}
  \begin{split}
  \left\{
    \begin{array}{c}    
    \begin{array}{rl}
      x_{k+1} &= G\left(x_k, w_k\right) \qquad x_0 = \mathtt{x}_0,\\
    	y_k 	&= h\left(x_k\right) + v_k,
      \end{array}
    \end{array}\right.
  \end{split}
\end{equation} 
\noindent where $x_{k} \in \mathcal{X} \subset \mathbb{R}^{n_x}$, $w_{k} \in \mathcal{W} \subset \mathbb{R}^{n_w}$, $y_k \in \mathcal{Y} \subset \mathbb{R}^{n_y}$ and $v_{k} \in \mathcal{V} \subset \mathbb{R}^{n_{v}}$ are the state, process disturbance, output, and measurement disturbance vectors, respectively. The process disturbance $w_k$ is assumed to be unknown but bounded, while the measurement noise $v_k$ is assumed to follow a specific prior distribution, though its exact form is not required. Furthermore, although noise is assumed to be bounded, some components of the vector $v_k$ occasionally have several samples with significantly higher values (\textit{outliers}). The sets $\mathcal{X}, \mathcal{W}, \mathcal{Y} \textnormal{ and } \mathcal{V}$ are compact and convex sets with the null vector $\mathbf{0}$ belonging to them. In the following, we assume that $G : \mathbb{R}^{n_x} \times \mathbb{R}^{n_u} \rightarrow \mathbb{R}^{n_x}$ is locally Lipschitz in its arguments and $h : \mathbb{R}^{n_x} \rightarrow \mathbb{R}^{n_y}$ is continuous.

\subsection{Adaptive Robust Loss Functions}
Baron \cite{Barron_2019_CVPR} introduced an adaptive robust loss that unifies several robust loss functions given by
\begin{equation}        \label{eq: general loss function}
  \rho\left(r,\alpha,c\right) = \frac{\vert \alpha-2\vert}{\alpha}\left(\left(1+\frac{1}{\vert\alpha-2\vert} \left(\frac{r}{c}\right)^{2}\right)^{\frac{\alpha}{2}-1}\right),
\end{equation}
\noindent where $r \in \mathbb{R}$ is the input, $\alpha \in (-\infty,2)$ is the shape-controlling parameter that yields different loss behaviours, and $c \in  \mathbb{R}_{>0}$ is a scale parameter defining the width of the quadratic region around $r=0$. Adjusting $\alpha$, different robust loss functions are generated. Some special cases are $L_2$ ($\alpha \to 2$), pseudo Hubber/$L_1-L_2$ ($\alpha=1$), Cauchy ($\alpha=0$), German-Mc Cure ($\alpha=-2$) and Welsch ($\alpha \rightarrow -\infty$) loss functions.
\begin{figure}[b]
 \centering
   \includegraphics[width=\linewidth]{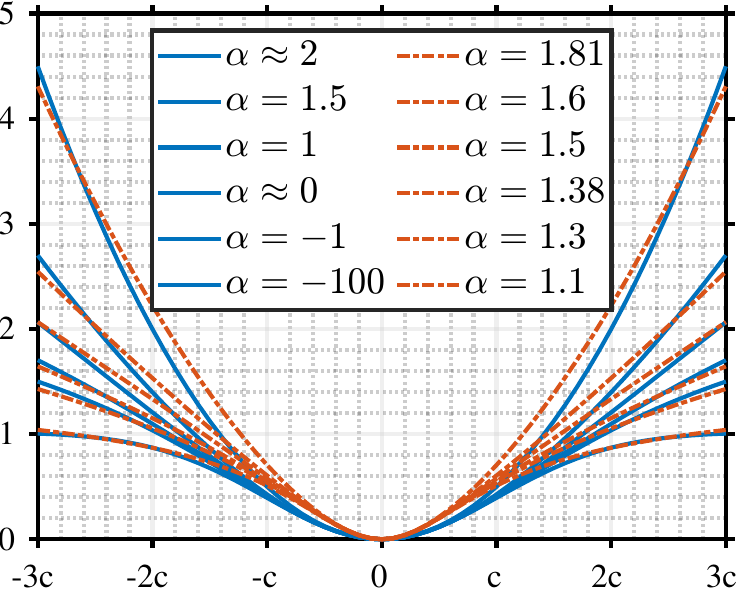}%
   \caption{Adaptive robust loss functions $\rho(r,\alpha,c)$ and $\varphi(r,\alpha,c)$ for different values of $\alpha$.}      \label{fig: garl and its derivative}
\end{figure}

Based on robust loss function \eqref{eq: general loss function}, Barron \cite{Barron_2019_CVPR} and Chebrolu et al. \cite{chebrolu2021adaptive} proposed an adaptive robust estimation framework where $\alpha$ is updated online. % via a grid search using a probabilistic approach. 
The main problems of using $\rho\left(r,\alpha,c\right)$ in a robust estimation framework are: (\textit{i}) its unbounded domain ($\alpha \in (-\infty,2)$) and (\textit{ii}) there is a discontinuity ($\alpha = 0$), which restrains the use of gradient-descendent optimization algorithms.

To overcome these problems, Deniz et al. \cite{deniz2025robust} introduced a new adaptive loss function built up from the square of the derivative of $\rho \left(r,\alpha,c \right)$ with respect to $r$, given by
\begin{equation}        \label{eq: derivative general loss function}
  \varphi \left(r,\alpha,c\right) = \frac{1}{c^2} \left(\frac{r}{c}\right)^{2}\left(1+\frac{1}{\vert\alpha-2\vert}\left(\frac{r}{c}\right)^{2}\right)^{\alpha-2}.
\end{equation}
\noindent Its behaviour is similar to $\rho\left(r,\alpha,c\right)$ for a bounded shape-controlling parameter domain ($\alpha \in (1,2)$) and has no discontinuities on the interval. Figure \ref{fig: garl and its derivative} shows the behaviours of $\rho \left(r,\alpha, c\right)$ and $\varphi \left(\hat{v}_{j\vert k},\alpha,c\right)$ for different values of the shape-controlling parameters $\alpha$.

Loss function $\varphi \left(r,\alpha,c\right)$ has the additional benefits of belonging to the class of $\mathcal{K}_{\infty}$ functions.
% , and it is convex $\forall \alpha\in(1,2)$. 
These kinds of functions are well-suited to design stage costs in moving-horizon estimators, since robust asymptotic stability can then be ensured under common and practical assumptions \cite{https://doi.org/10.1049/iet-cta.2019.0523,https://doi.org/10.1002/rnc.6799}.

\subsection{Outliers-robust $\textrm{MHE}$}
Let us start by formulating a robust $\textrm{MHE}$ that accurately estimates the states of system \eqref{eq: eq_nonlinsys} from measurements with outliers. In a first step, the stage cost of the estimation problem $\ell \left(\hat{w}_{j\vert k}, \hat{v}_{j\vert k}\right)$ is modified to include the adaptive loss function and a regularization term for $\alpha$
\begin{equation}
  \begin{split}
    \ell \left(\hat{w}_{j},\hat{v}_{j},\alpha\right) & = \ell_{w}\left(\hat{w}_{j}\right) + \frac{1}{n_y}\sum^{n_y}_{i=1} \delta_i \, \varphi\left(\hat{v}_{i,j},\alpha,c\right)\\
   & \quad+ \,\gamma \, \left(2-\alpha_{k}\right)^{2} \quad \delta_i \in (0,1], \gamma \in \mathbb{R}_{>0},
  \end{split}
\end{equation}
where $j \in \mathbb{Z}_{[k-N,k]}$ and $\hat{v}_{i,j}$ is $i$-\textit{th} component of the $j$-\textit{th} estimation residuals. The term $\ell_{w}\left(\hat{w}_{j}\right)$ penalizes $\hat{w}_{j}$, acting as a regularization term of $\hat{\boldsymbol{x}}_{j}$, $\varphi(\cdot)$ is the adaptive loss function defined in Eq. \eqref{eq: derivative general loss function} that weights the residuals $\hat{v}_{j}$ and the third term penalizes the inflation of the outliers set (controlled by $\alpha$) by measuring the deviation of $\varphi \left( \cdot \right)$ from an $L_2$ loss function. It acts as a regularization term for the shape-controlling parameter $\alpha$ by maximizing its value. 
%%%%%%%% In this way, the optimization problem searches for solutions that minimize a measure of the residuals $\hat{\boldsymbol{v}}_{[k-N,\,k]}$ while making $\varphi \left( \cdot \right)$ as close as possible to an $\textrm{L_2}$. It introduces a trade-off between the cost of penalizing fewer outliers and the increasing value of the shape-controlling parameter, and vice versa. This trade-off forces the optimization to choose a suitable value of $\alpha$ instead of trivially ignoring all residuals by turning every measurement into an outlier. %%%%%%%
%This work formulates a real-time MHE to accurately estimate the system states governed by Eq. \eqref{eq::eq_nonlinsys}. 
%Instead of the conventional $\mathrm{L_2}$ stage cost, we employ the adaptive robust loss function
%\begin{equation}        \label{eq::charbonnier}
%  \begin{array}{rl}
%    \psi(x,\alpha,c) = \frac{r^2}{c^4}\left(\frac{r^2} {c^2\vert\alpha-2\vert}+1\right)^{\alpha-2},
%  \end{array}
%\end{equation}
Then, the objective function for the outlier-robust $\textrm{MHE}$ estimator is given by
\begin{equation}        \label{eq: ct objective function mhe}
  \begin{array}{rl}
    \mathbb{J} %\left(\hat{\boldsymbol{w}},\hat{\boldsymbol{v}},\alpha_{k}\right)
    =& \Gamma_{k-N\vert k} \left(\chi\right) + \sum\limits_{j=k-N}^{k} \ell_{w}\left(\hat{w}_{j\vert k}\right) \\
    & +  \frac{1}{n_{y}}\sum\limits_{i=1}^{n_y} \delta_i \, \varphi \left(\hat{v}_{i,j\vert k},\alpha_{k},c\right) + \, \gamma \, \left(2-\alpha_{k}\right)^2,   
  \end{array}
\end{equation}
\noindent where is the arrival residual $\hat{\chi} = \hat{x}_{k-N\vert k}-\bar{x}_{k-N\vert k}$ is updated according to \cite{sanchez2017adaptive}. %, and $\hat{v}_{i,j}$ is $i$-\textit{th} component of the estimation residuals
Then, the outlier-robust $\textrm{MHE}$ estimator is given by
\begin{equation}                        \label{or_sa_mhe}
 \begin{array}{cl}
\mathbb{J}^{*} =& \underset{\hat{x}_{k-N\vert k},\boldsymbol{\hat{w}}_{k},\alpha_{k}} {\operatorname{argmin}} \;\mathbb{J} \\ 
  \text{s.t.} & \left\{
  \begin{array}{l}
    \begin{array}{rlc}
      \chi =& \hat{x}_{k-N\vert k} - \bar{x}_{k-N \vert k}, \vspace{1mm}\\
      \hat{x}_{j+1\vert k} =& G\left(\hat{x}_{j\vert k}, \hat{w}_{j\vert k}\right),\vspace{1mm} \\
    y_{j} =& h\left(\hat{x}_{j\vert k}\right) + \hat{v}_{j\vert k}, \vspace{1mm}\\
    \end{array}                  \\
    \hat{x}_{j|k} \in \mathcal{X}, \, \hat{w}_{j|k} \in \mathcal{W}, \\ \hat{v}_{j|k} \in \mathcal{V}, \alpha_{k} \in \left(1,2\right).
  \end{array} \right.
 \end{array}
\end{equation}
%\noindent In the second step, we modify the arrival cost updating mechanism as follows
%\begin{equation}
%  \begin{split}
%    N_k = &\left[ 1 + \hat{x}_{k-N\vert k-1}^T\, P_{k-N-1\vert k-1} \hat{x}_{k-N\vert           k-1}\right] \vspace{0.1cm} \\
%          & \quad\frac{\sigma}{\sum\limits_{i=1}^{n_y} \varphi \left(\hat{v}_{i,k-N\vert k},\alpha_{\varphi},c\right)},  
%  \end{split}
% \end{equation}
%\noindent to attenuate the effects of outliers on $P_{k-N\vert k}$. 
Optimization problem \eqref{or_sa_mhe}, combined with the arrival cost updating mechanism, estimates a shape-controlling parameter $\alpha_{k}$ that enhances the robustness of the $\textrm{MHE}$ estimation problem by mitigating the effects of outliers within the estimation window by searching for solutions that minimize a measure of the residuals $\hat{\boldsymbol{v}}_{[k-N,\,k]}$ while making $\varphi \left( \cdot \right)$ as close as possible to an $\mathrm{L_2}$. It introduces a trade-off between the cost of penalizing fewer outliers and the increasing value of the shape-controlling parameter, and vice versa. This trade-off forces the optimization to choose a suitable value of $\alpha_{k}$ instead of trivially ignoring all residuals by turning every measurement into an outlier. 

This approach limits the estimator's ability to adapt to the underlying probability distribution of outliers since it assumes a homogeneous distribution across the estimation window. It simplifies the estimation process but may overlook data heterogeneity \cite{em2011robust}, leading to a reduction of robustness in scenarios where outliers are localized or temporally clustered \cite{schiller2024moving}. However, from a regularization perspective, constraining the residuals of an estimation window to share a common robustness level acts as an implicit structural constraint that can enhance estimation stability in scenarios with limited excitation \cite{baumgartner2022moving}.

Assigning a separate shape-controlling parameter to each measurement $y_{k}$ provides a higher degree of adaptability, allowing the estimator to respond differentially to outliers at specific points in the horizon, thereby improving robustness against localized anomalies. This idea leads to the following stage cost 
\begin{equation}
\label{eq::full opt problem}
  \begin{split}
    \ell \left(\hat{w}_{j},\hat{v}_{j},\alpha_{j}\right) & = \ell_{w}\left(\hat{w}_{j}\right) + \frac{1}{n_y}\sum^{n_y}_{i=1} \delta_i \, \left(\varphi\left(\hat{v}_{i,j},\alpha_{i,j},c\right) \right.\\
   & \left. \quad+ \, \gamma_{i} \,\left(2-\alpha_{i,j}\right)^{2}\right),
  \end{split}
\end{equation}
\noindent and the outlier-robust $\textrm{MHE}$ estimator is given by
\begin{equation}                        \label{or_ma_mhe}
 \begin{array}{cl}
\mathbb{J}^* =& \underset{\hat{x}_{k-N|k},\boldsymbol{\hat{w}}_{k},\boldsymbol{\alpha}_{k}} {\operatorname{argmin}} \;\mathbb{J} \\ 
  \text{s.t.} & \left\{
  \begin{array}{l}
    \begin{array}{rlc}
      \chi =& \hat{x}_{k-N\vert k} - \bar{x}_{k-N \vert k}, \vspace{1mm}\\
      \hat{x}_{j+1\vert k} =& G\left(\hat{x}_{j\vert k}, \hat{w}_{j\vert k}\right),\vspace{1mm} \\
        y_{j} =& h\left(\hat{x}_{j\vert k}\right) + \hat{v}_{j\vert k}, \vspace{1mm}      \\
    \end{array}                  \\
    \hat{x}_{j\vert k} \in \mathcal{X}, \, \hat{w}_{j\vert k} \in \mathcal{W}, \\ \hat{v}_{j\vert k} \in \mathcal{V}, \alpha_{i,j\vert k} \in \left(1,2\right).
  \end{array} \right.
 \end{array}
\end{equation}
This formulation enables the estimator to adaptively determine the degree of robustness for each measurement within the estimation horizon, thereby accommodating heterogeneous outlier distributions and time-varying measurement quality. The adaptive mechanism automatically balances the trade-off between fitting inliers and rejecting outliers at each time instant by treating shape-controlling parameters as latent variables whose values are optimized to maximize the likelihood of the observed data. However, this increment of flexibility comes at computational cost, introducing heightened risk of overfitting, particularly in scenarios with sparse measurements or limited horizon lengths, where the number of shape parameters may approach or exceed the number of measurements.

\subsection{Algorithm implementation}     \label{sec: two-stg optimization}
Optimization problems \eqref{or_sa_mhe} and \eqref{or_ma_mhe} aim to estimate simultaneously the state trajectory, the associated variables, and the shape-control parameters that minimizes the effects of the outliers present in the estimation window $\boldsymbol{y}_{[k-N,k]}$. However, the product of the optimization variables impose a challenge. To overcome this disadvantage, the optimization problem \eqref{or_ma_mhe} is recast as a dual estimation problem in which the state trajectory and shape-control parameter are estimated separately within a sequential approach where a state estimation problem is solved assuming a fixed $\boldsymbol{\alpha}_{k}$,
%\begin{equation}        \label{eq::discrete-time OF}
%  \begin{array}{rl}
%    \mathbb{J} =& \Gamma_{\chi}(\hat{\chi}) + \displaystyle\sum_{\substack{j=k-N_e \\ l=k-N_e}}^{\substack{k\\k-1}}\left( \ell_{w} ( \hat{w}_{l\vert k}) + \sum_{i=1}^{n_r}\psi \left( \hat{\nu}_{i,j\vert k},\alpha_{i,j\vert k},c\right)\right).
%  \end{array}
%\end{equation}
%\noindent Then, the optimization problems to be solved at each sampling time are
\begin{equation}        \label{eq:discrete-time-Jx}
  \begin{array}{c}
    \mathbb{J}_x:=\underset{\hat{x}_{k-N|k},\boldsymbol{\hat{w}}_{k}}{\operatorname{argmin}}\quad\mathbb{J} \vspace{1.5mm}\\
    \text{s.t.} \left\{
    \begin{array}{l}
      \begin{array}{rl}
        \hat{\chi} =& \hat{x}_{k-N_e\vert k} - \bar{x}_{k-N_e\vert k}, \vspace{1.5mm}\\
        \hat{x}_{j+1} =& G\left(\hat{x}_{j\vert k},\hat{w}_{j\vert k}\right)+\hat{w}_{j\vert k},\vspace{1.5mm}  \\
        r_{j} =& h\left(\hat{x}_{j\vert k}\right) + \hat{v}_{j\vert k},\vspace{1.5mm} \\
      \end{array} \vspace{1mm} \\
      \;\hat{x}_{j\vert k}\in\mathcal{X},\hat{w}_{j\vert k}\in\mathcal{W},\,\hat{v}_{j\vert k}\in\mathcal{V},
   \end{array} \right.
  \end{array}
\end{equation}
\noindent and
\begin{equation}        \label{eq:discrete-time-J-alpha}
  \begin{array}{c}
    \mathbb{J}_{\alpha}:=\underset{\boldsymbol{\alpha}_{k}}{\operatorname{argmin}}\quad\mathbb{J} \vspace{1.5mm}\\
    \text{s.t.} \left\{
      \begin{array}{l}
        \begin{array}{c}
          1<\alpha_{i,j\vert k}<2.
      \end{array}
    \end{array} \right.
  \end{array}
\end{equation}
\noindent with $\boldsymbol{\hat{x}}_{[k-N,k]}$, $\boldsymbol{\hat{w}}_{[k-N,k-1]}$ and $\boldsymbol{\hat{v}}_{[k-N,k]}$ remaining fixed.
Solving iteratively problems \eqref{eq:discrete-time-Jx} and \eqref{eq:discrete-time-J-alpha} generates a non-increasing sequence of costs $\mathbb{J}_{\hat{\boldsymbol{x}}_1}$, $\mathbb{J}_{\boldsymbol{\alpha}_1}$, ... , $\mathbb{J}_{\hat{\boldsymbol{x}}_\mathrm{M}}$, $\mathbb{J}_{\boldsymbol{\alpha}_\mathrm{M}}$ and the estimation process is stopped when $\mathbb{J}_{\hat{\boldsymbol{x}}_i}-\mathbb{J}_{\boldsymbol{\alpha}_i} \leq \varepsilon$ with $1\leq i \leq \mathrm{M}$ and some $\varepsilon>0$, or $i$ reaches the predefined maximum number of iterations $\mathrm{M}$. This sequence of costs is lower-bounded, non-increasing. Thus, it is convergent. A proof can be sketched following the steps described in \cite{ekeland1979nonconvex}. The algorithm \ref{alg::alg iterative estimation} summarises the iterative estimation procedure.

This sequence of costs is generated at each sampling time until it reaches the stop condition. At the next sampling instant ${k+1}$, the values of $\boldsymbol{\alpha}_{k}$ are initialized (receded) as $\alpha_{i,j-1\vert k+1}=\alpha_{i,j\vert k}$ while $\alpha_{i,k+1\vert k+1} \rightarrow 2$, i.e. the values of $\boldsymbol{\alpha}_{k+1}$ associated with the last measurement incorporated into the estimator are initialized such that the stage cost corresponds to a standard $\mathrm{L}_2$ behaviour.
\begin{algorithm}[tb]    \caption{Iterative state and shape parameter estimations}
\label{alg::alg iterative estimation}
\KwData{$\bar{x}_{k-\mathrm{N_e}},\,\Phi,\,h,\,\mathcal{W},\,\mathcal{V}$.}
\KwResult{$\hat{\boldsymbol{x}}_{k-N \vert k},\,\boldsymbol{\alpha}_{i,[k-N\vert k]}$}
\textbf{Initialise:} $\alpha_{i,[1,\mathrm{N}]}\rightarrow 2$, $\varepsilon>0$\\
\While{True}{
$i=1$\\
$\Delta\mathbb{J}\rightarrow\infty$\\
\While{$\Delta\mathbb{J}>\varepsilon \;\mathrm{AND}\; i\leq\mathrm{M}$}{
$\mathbb{J}_{x_i}\gets$ Solve Eq. \eqref{eq:discrete-time-Jx}\\
$\mathbb{J}_{\alpha_i}\gets$ Solve Eq. \eqref{eq:discrete-time-J-alpha}\\
$\Delta\mathbb{J}=\mathbb{J}_{\hat{\boldsymbol{x}}_i}-\mathbb{J}_{\boldsymbol{\alpha}_i}$\\
$i=i+1$}
$\bar{x}_{k-\mathrm{N}}=\hat{x}_{k-\mathrm{N}+1\vert k}$\\
$\alpha_{i,[{k}-\mathrm{N}+1,{k}]\vert {k+1}}=\alpha_{i,[{k}-\mathrm{N}+1,{k}]\vert k}$\\
$\alpha_{i,{k}\vert {k+1}}\rightarrow2$\\
Update measurements}
\end{algorithm}
\vspace{5cm}

\subsection{Robust stability}
\noindent The robust stability of the estimator given by \eqref{eq:discrete-time-Jx}-\eqref{eq:discrete-time-J-alpha} is guaranteed under standard MHE assumptions: (i) system detectability, (ii) continuity of the stage cost, and (iii) bounded arrival and stage costs.  Then, let us assume that the system \eqref{eq: eq_nonlinsys} is detectable. The continuity of the stage cost is guaranteed since both $\ell_w$ and $\varphi(\hat{v},\alpha,c)$ are continuous functions. Specifically, $\ell_w$ is the standard $\mathrm{L_2}$ loss, and $\varphi(\hat{v},\alpha,c)$, defined in \eqref{eq: derivative general loss function} is smooth, continuous, and differentiable $\forall x, c > 0 \text{ and } \alpha\in[1,2)$.

\noindent The lower and upper bounds of the stage cost also hold. The term $\ell_w(\hat{w}_{\tau})$ can be expressed as $\hat{w}_{\tau}^T W \hat{w}_{\tau}$, with $W$ a positive definite matrix. Hence, the inequality $\underline{\lambda}_W | \hat{w}_{\tau} |^2 \leq \ell_w(\hat{w}_{\tau}) \leq \overline{\lambda}_W | \hat{w}_{\tau} |^2$ holds, where $\underline{\lambda}_W$ and $\overline{\lambda}_W$ are the minimum and maximum eigenvalues of $W$. Similarly, the arrival cost satisfies $\underline{\lambda}_{\Gamma} | \hat{\chi} |^2 \leq \Gamma(\hat{\chi}) \leq \overline{\lambda}_{\Gamma} | \hat{\chi} |^2$ (see \cite{sanchez2017adaptive} for details). Furthermore, $\varphi(\hat{v},\alpha,c)$ can be bounded as $\varphi(\hat{v},\alpha-\epsilon,c) \leq \varphi(\hat{v},\alpha,c) \leq \varphi(\hat{v},\alpha+\epsilon,c)$ for a small $\epsilon > 0$ and constraining $\alpha > 1$.

\noindent The key to proving the robust stability of the estimator \eqref{eq:discrete-time-Jx}-\eqref{eq:discrete-time-J-alpha} lies in bounding the estimated variables in terms of the true variables. Let $\mathbb{J}_{\hat{\boldsymbol{x}}}^{*}$ denote the optimal cost, defined as
\begin{equation*}
  \begin{array}{rl}
    \mathbb{J}_{\hat{\boldsymbol{x}}}^{*} =& \Gamma_{\chi}\left(\hat{\chi}^*\right) + \sum_{\substack{j=k-N \\ l=k-N}}^{\substack{k\\k-1}}\left( \ell_{w}\left(\hat{w}_{l\vert k}^*\right) \right. \vspace{1mm}\\
    &\qquad \left. +\sum_{i=1}^{n_y}\varphi \left( \hat{v}_{i,j\vert k}^*,\alpha_{i,j\vert k}^*,c\right)\right), \\
%  \end{array}
%\end{equation*}
%\begin{equation*}
%  \begin{array}{rl}
    \leq& \Gamma_{\chi}\left(\chi\right) + \sum_{\substack{j=k-N \\ l=k-N}}^{\substack{k\\k-1}}\left( \ell_{w} ( w_{l \vert k}) \right.\vspace{1mm}\\
    &\qquad \left. + \sum_{i=1}^{n_r}\varphi \left( \hat{v}_{i,j\vert k},\alpha_{i,j\vert k}^*,c\right)\right),
  \end{array}
\end{equation*}
\noindent where $\chi$, $w_j$, and $\hat{v}_j$ denote the true arrival cost, process disturbance, and measurement residual, respectively. The inequality follows from the optimality of the estimator, enabling a link between the estimated and true variables. Using the previously established bounds, the optimal cost $\mathbb{J}^*$ can be upper-bounded as:
\begin{equation*}
  \mathbb{J}_q^* \leq \overline{\lambda}_{\Gamma}\vert\chi\vert^2 + \mathrm{N}\,\overline{\lambda}_W\Vert \mathbf{w} \Vert^2 + (\mathrm{N_e}+1)f(\Vert\boldsymbol{\nu}\Vert,k_{\alpha}+\epsilon,c),
\end{equation*}
\noindent where $\Vert\mathbf{w}\Vert = \max\{|w_i|\}$, $\Vert\boldsymbol{\hat{v}}\Vert = \max\{\norm{\hat{v}_{i}}\}\, \forall i \geq 0$. Following a similar algebraic procedure as in \cite{https://doi.org/10.1049/iet-cta.2019.0523}, and selecting the window length $\mathrm{N} \geq \mathbb{N}_{\min}$ -with $\mathbb{N}_{\min}$ defined in Eq. (58) of \cite{https://doi.org/10.1049/iet-cta.2019.0523}— the estimator can be shown to be robustly asymptotically stable (RAS). %$\square$

\section{Experiments and Results}       \label{sec: experiments and results}
\noindent In this section, we evaluate the performance of the proposed scheme ($\mathrm{MHE_{prop}}$) and compare it against our former method with fixed $\alpha$ and the grid-search method for adapting $\alpha$ over a set of predefined values. The evaluation involves simulated experiments for a path-following task, where a tractor pulls a single passive trailer, with GNSS measurements affected by outliers. The path-following task requires the system to track a path commonly encountered in agricultural environments.

The 1-trailer vehicle consists of a differentially driven tractor \cite{DENIZ2024105747} pulling a passive trailer, connected through a rotary joint. The system state is given by $q_k = (\beta_{1,k}, \theta_{0,k}, \theta_{1,k}, x_{0,k}, y_{0,k}, x_{1,k}, y_{1,k})^T$, where $\theta_{0,k}$ and $\theta_{1,k}$ denote the orientations of the tractor and trailer, and $\beta_{1,k} = \theta_{0,k} - \theta_{1,k}$ is the joint angle between the tractor and the trailer. The vehicle positions are given by $(x_{0,k}, y_{0,k})$ for the tractor and $(x_{1,k}, y_{1,k})$ for the trailer. The control vector is $u_k = (\omega_{0,k}, v_{0,k})^T$, representing the tractor's steering rate and velocity. The kinematic model is compactly written as ${q}_{k+1} = \mathbb{G}(q_k) u_k$. The measurement vector $r_k$ is defined as:
\begin{align}        
    r_k = h(q_k) = (\beta_{1,k}, \theta_{0,k}, x_{0,k},y_{0,k})^T, \nonumber
\end{align}
\noindent where $h:\mathbb{R}^{n_q}\rightarrow\mathbb{R}^{n_r}$, is the output function. Thus, the 1-trailer vehicle is succinctly described by the following equations:
\begin{equation}        \label{eq::continuous-time system}
  \left\{
    \begin{array}{c}    
      \begin{array}{rl}
        {q}_{k+1} =& \mathbb{G}(q_k)\,u_k + w_k, \vspace{1.5mm}\\
              r_k =& h(q_k) + \nu_k,
      \end{array}
    \end{array}
  \right.
\end{equation}
\noindent where vector $w_k\in\mathbb{R}^{n_q}$ models any additive process, or model disturbance, while vector $\nu_k\in\mathbb{R}^{n_r}$ is the additive measurement noise. The measurement noise is assumed to follow a specific prior distribution, though its exact form is not required. In spite of assuming a bounded measurement noise, the tractor coordinates measurements $(x_{0,k},\,y_{0,k})$ occasionally exhibit larger values (outliers). The vehicle is controlled using the model predictive controller (MPC) presented in \cite{DENIZ2024105747} where both speed and acceleration are constrained to achieve smooth movements.

Two scenarios are considered: In the first scenario measurement noise following a normal distribution with zero mean and variances $\sigma_{\beta}^2 = (\pi/180)^2$, $\sigma_{\theta}^2 = (0.2\pi/180)^2$, and $\sigma_{xy}^2 = (0.025)^2$, and (ii) measurement noise following a uniform distribution with supports $\pm\sigma_{\beta}$, $\pm\sigma_{\theta}$ and $\pm\sigma_{xy}$ on the joint angle, attitude and position of the tractor, respectively. In both scenarios, outliers are randomly distributed over the estimation window with appearance probability $\delta$ and $\sigma^2=100$ for scenario (i) and supports $\pm 10$ for the second scenario.

\subsection{Performance evaluation}
We compare the performance of the proposed estimator ($\mathrm{MHE_{prop}}$) configured with an estimation window length $\mathrm{N_e} = 10$,  $c = 1$, $\varepsilon=1e^{-3}$ and $\mathrm{M}=10$. 
Since no process disturbances are assumed to be present, the covariance matrix $W$ is configured as a diagonal matrix with large values: $W = 10^6\times\mathrm{I}$, where $\mathrm{I}$ is the identity matrix of appropriate dimensions. The initial guess for the estimator is generated by adding normal noise with a standard deviation of $\sigma = 3$ (m) to the true position of the tractor. From this position, a feasible configuration is generated for the chained vehicle. This large initial guess error is intentionally introduced to prevent the estimator from relying solely on the vehicle model and to force it to incorporate measurements. The sets $\mathcal{W}$ and $\mathcal{V}$ in Eq. \eqref{eq:discrete-time-Jx} are defined as box constraints. Each component $\hat{w}$ is limited to $-0.01 \leq \hat{w}_{i,k} \leq 0.01$. For $\mathcal{V}$, residual components not affected by outliers are bounded by $-0.1 \leq \hat{\nu}_{i,k} \leq 0.1$, while those corresponding to the measurements affected by outliers are unbounded.

\begin{figure}[!b]
\centering
\begin{minipage}{\linewidth}
    \includegraphics[width=\linewidth]{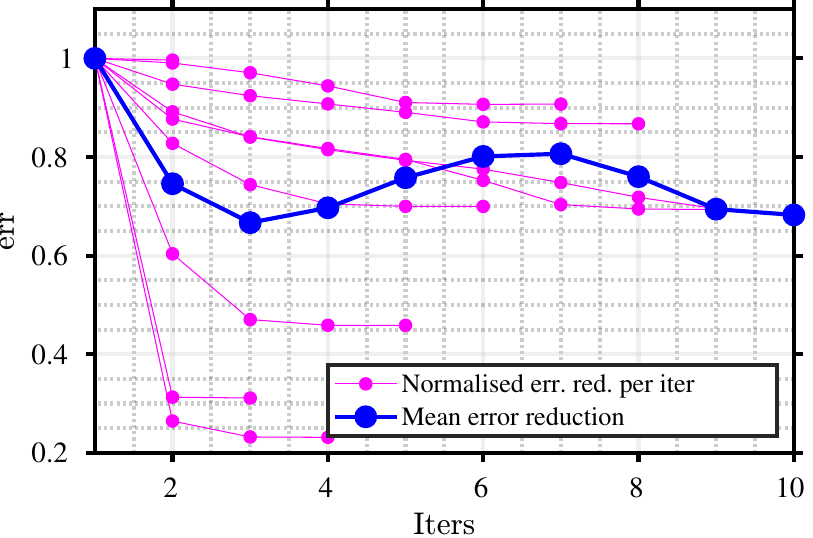}%
\end{minipage}\\
\begin{minipage}{\linewidth}
    \includegraphics[width=\linewidth]{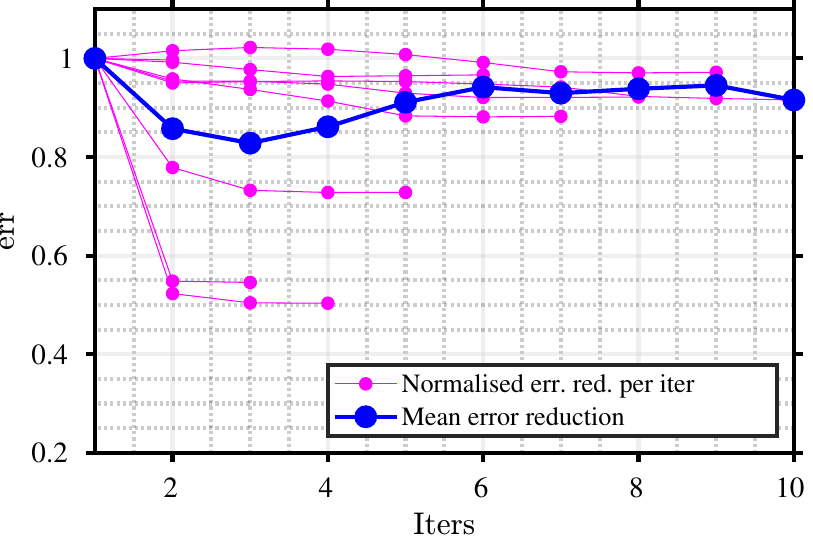}%
\end{minipage}\\
%\begin{minipage}{\linewidth}
%    \includegraphics[width=\linewidth]{Figs/times_adapt_normal.pdf}%
%\end{minipage}
\caption{Normalized estimation error reduction per iteration and mean estimation error reduction under scenario (i) (top) and scenario (ii) (middle).}% Total computational load as a function of the number of iterations performed.}
\label{fig::sim exps 1000 trials}
\end{figure}
Figure \ref{fig::sim exps 1000 trials}.\textit{a} shows the reduction in normalized estimation error, averaged over $1000$ trials in the first scenario, while Figure \ref{fig::sim exps 1000 trials}.\textit{b} shows the results in the second scenario. The blue lines indicate the mean error reduction per iteration. It should be noted that, at each sampling time, the estimator does not always achieve the maximum number of iterations $\mathrm{M}$. The iterative estimation process is stopped when $\Delta\mathbb{J} \leq \varepsilon$. As can be seen, the mean error reduction decreases during the iterative estimation process. However, the reduction does not continue beyond the third iteration. Although a higher error reduction is observed in the first scenario, the same behaviour is observed with respect to the mean error reduction in both scenarios.% The bottom of Figure \ref{fig::sim exps 1000 trials} shows the mean time required to compute the estimation as a function of the number of iterations at each sampling time. 

\begin{figure}[!b]
\centering
\begin{minipage}{\linewidth}
    \includegraphics[width=\linewidth]{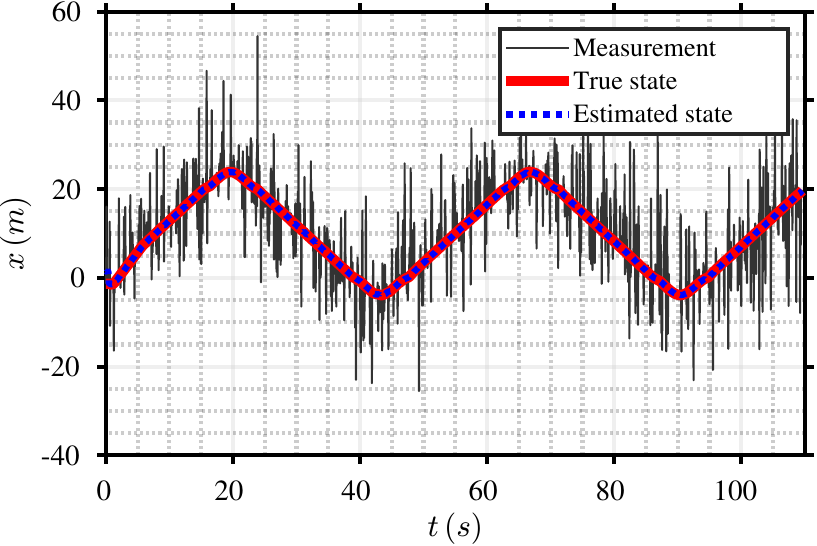}%
\end{minipage}\\
\begin{minipage}{\linewidth}
    \includegraphics[width=\linewidth]{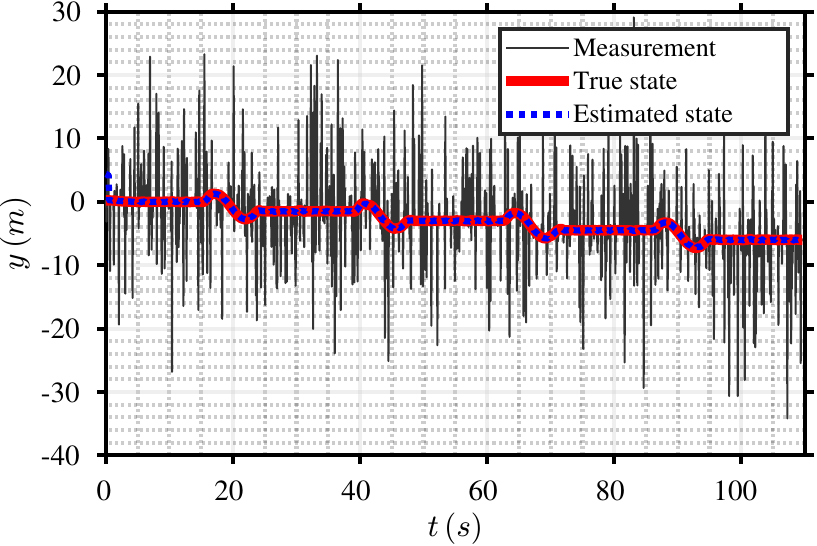}%
\end{minipage}\\
\begin{minipage}{\linewidth}
    \includegraphics[width=\linewidth]{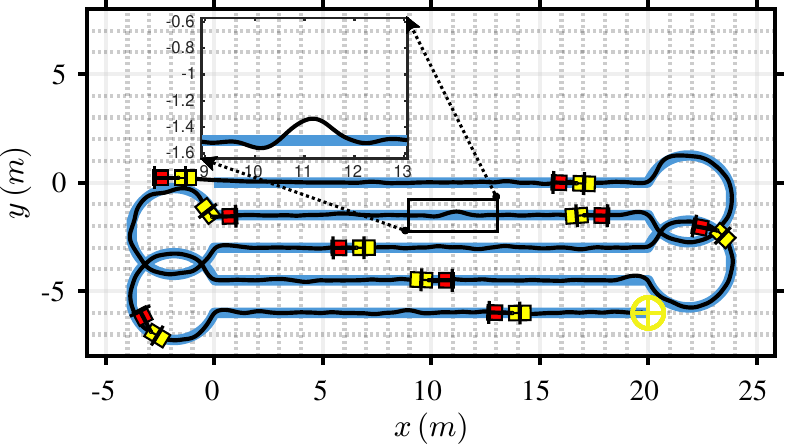}%
\end{minipage}
\caption{Measured, true, and estimated tractor's $x$ and $y$ coordinates for one trial of the experiments. The bottom panel illustrates the trajectory followed by a tractor pulling a trailer.}% (plotted at $9$ different times), showing a maximum deviation of approximately $0.25$ (m) from the path when $25$ \% of the tractor's position measurements are corrupted by outliers with amplitudes of up to $20$ (m).}
\label{fig::sim exp 1 trial}
\end{figure}
Figure \ref{fig::sim exp 1 trial} shows the tractor’s $\mathrm{x}$ and $\mathrm{y}$ coordinates for one trial, as well as the trajectory followed and the vehicle’s position at $9$ different times. Since the mean error reduction does not decrease after the third iteration, we set $\mathrm{M} = 3$ to compare with $\mathrm{MHE_{grid\,s.}}$ and $\mathrm{MHE_{\alpha, fixed}}$. In addition, $\mathrm{MHE_{grid\,s.}}$ is also configured with $\mathrm{M} = 3$, such that $\alpha \in {1.1,\,1.5,\,1.8}$. Moreover, all estimators are configured with an estimation window length $\mathrm{N_e} = 10$ and $c = 1$. The experiments aim to evaluate performance in terms of the following metrics: the mean squared estimation error ($\Psi$), the mean time ($\eta$) required to compute it, and the mean squared tractor position error ($\Delta$). These metrics are computed as follows
\begin{equation*}
    \begin{array}{rl}
        \Psi =& \frac{1}{\mathrm{N}}\sum_{i=1}^{\mathrm{N}}(\hat{q}_{k|k}-q_k)'\,(\hat{q}_{k|k}-q_k),\vspace{1.5mm}\\
        \eta =& \frac{1}{\mathrm{N}}\sum_{i=1}^{\mathrm{N}}t_i,\vspace{1.5mm}\\
        \Delta =& \frac{1}{\mathrm{N}}\sum_{i=1}^{\mathrm{N}}(\hat{x}_{0,\,k\vert k}-x_{0,k})^2+(\hat{y}_{0,\,k\vert k}-y_{0,k})^2,
    \end{array}
\end{equation*}
\noindent where $(x_{ref,k},\,y_{ref,k})$ is the reference position to be followed at time $k$; $(x_k,\,y_k)$ is the actual position of the tractor at time $k$; $(\hat{x}_{k|k},\,\hat{y}_{k|k})$ is the estimated position of the tractor at time $k$; $t_i$ is the time required to solve the estimation problem at instant $i$, and $\mathrm{T_s}$ is the sampling time.
\begin{table*}[thb]
\centering
\caption{Comparison of MHE methods under different noise distributions computed over $1000$ trials}
\begin{tabular}{lcccccc}
\toprule
\textbf{Method} 
& \multicolumn{3}{c}{\textbf{Uniform Noise}} 
& \multicolumn{3}{c}{\textbf{Normal Noise}} \\
\cmidrule(lr){2-4} \cmidrule(lr){5-7}
& MS. Est. Err. ($\Psi$) & MS. Pos. Err. ($\Delta$) & Mean Time ($\eta$)
& MS. Est. Err. ($\Psi$) & MS. Pos. Err. ($\Delta$) & Mean Time ($\eta$) \\
\midrule
$\text{MHE}_{\text{prop}}$ (M=10) & \textbf{0.1708} & \textbf{0.1154} & 0.3711 & \textbf{0.1644} & 0.1106 & 0.3744 \\
$\text{MHE}_{\text{prop}}$ (M=3)  & 0.1711 & \textbf{0.1154} & 0.1090 & 0.1649 & 0.1106 & 0.1144 \\
$\text{MHE}_{\text{grid s.}}$ (M=3) & 0.1885 & 0.1284 & 0.0732 & 0.1624 & \textbf{0.1093} & 0.0733 \\
$\text{MHE}_{\alpha\ \text{fixed}}$ & 0.1944 & 0.1326 & \textbf{0.0290} & 0.1962 & 0.1337 & \textbf{0.0307} \\
\bottomrule
\end{tabular}
\label{tab::MHE_comparison}
\end{table*}

Table \ref{tab::MHE_comparison} summarizes the results obtained for the three variants compared. As expected, $\mathrm{MHE_{\alpha\, fixed}}$ achieves the lowest computational load at the expense of estimation error, although its performance remains very good, as reported in our previous work \cite{deniz2025robust}. $\mathrm{MHE_{grid\,s.}}$ achieves a good trade-off between the estimation error and the computational load. It should be noted that the performance of $\mathrm{MHE_{\alpha\, fixed}}$ is covered by $\mathrm{MHE_{grid\,s.}}$, since $\alpha = 1.5$ is included among the search values. Finally, the performance of $\mathrm{MHE_{prop}}$ appears to be only marginally better, at the cost of a higher computational load. One point in its favour is that the distribution of the estimated values of $\alpha$ over time could be useful for characterizing how noise affects the measurements. We will address this point in future work.

\section{Conclusions and Future Works}      \label{sec: conclusions}
In this work, we proposed a Moving Horizon Estimator that replaces the traditional $\mathrm{L_2}$ stage cost with an adaptive robust loss defined as the square of the derivative of Barron’s General Adaptive Robust Loss \cite{Barron_2019_CVPR}. This design improves estimation accuracy under severely unfavourable conditions, but the performance gain comes at increased computational cost relative to both our previous method and a variant that tunes $\alpha$ via grid search. Future work will reformulate the optimization problem in \eqref{eq::full opt problem} to eliminate products of decision variables and the dual-estimation scheme, with the goal of maintaining high accuracy while reducing computational cost.

\section*{Acknowledgments}
The authors thank the Consejo Nacional de Investigaciones Científicas y Técnicas (CONICET) from Argentina. 
Since the authors are not native English speakers, the text was revised and improved with chatGPT-5.

\bibliographystyle{IEEEtran}
\bibliography{biblio.bib} 
% \end{thebibliography}b

% \bibliographystyle{agsm}% Include this if you use bibtex 
% \clearpage
% \bibliography{autosam.bib} 

\newpage

 

% \end{IEEEbiography}



\vfill

\end{document}